\documentclass[10pt,twocolumn]{article}
\usepackage[margin=0.75in]{geometry}
\usepackage{times,amsmath,amssymb,booktabs,graphicx,xcolor,caption}
\usepackage[hidelinks]{hyperref}
\usepackage{placeins}
\title{Using Action-Conditioned Lookahead in a Learned Game Policy:\\A Tank Trouble Case Study}
\author{Weiyi Jiang\\Australian National University\\Canberra, ACT, Australia\\\small\texttt{cichlid1234@outlook.com}}
\date{}
\hypersetup{pdftitle={Using Action-Conditioned Lookahead in a Learned Game Policy: A Tank Trouble Case Study},pdfauthor={Weiyi Jiang}}
\begin{document}
\raggedbottom
\maketitle
\begin{abstract}
Short-horizon simulation can provide a game policy with action-dependent predictions,
but those predictions can enter the policy in different ways. We study three routes:
reading lookahead scores as neural inputs, adding them to action logits, and using
them only as auxiliary training targets. In Tank Trouble, we evaluate these routes
in Hybrid, an augmented PPO actor--critic, against a heuristic opponent and an MPC
opponent. The variants continue from a common pretrained checkpoint with matched
interaction budgets. Across three continuation seeds, shared actor--critic access to
the scores yields the largest observed improvement over the no-dodge baseline.
Additional scalar logit biases have smaller, uncertain gains. A training-only
supervision implementation also has three completed runs, but does not recover the
input route's performance. Same-state interventions reveal limited immediate
sensitivity to the direct bias in an exploratory state bank. We distinguish online
lookahead requirements from measured deployment speedups: no matched end-to-end
latency ranking is established here. This case study illustrates why access to a useful
signal, its immediate influence on decisions, and its task benefit should be assessed
separately.
\end{abstract}

\section{Introduction}
A short simulation can identify movements that avoid an incoming projectile without
determining which movement best serves a duel objective. A learned policy can
interpret these predictions alongside other observations, accept them as a direct
action bias, or learn from them during training. These choices differ in how they
shape learning and whether deployment still requires the simulation.

\begin{figure*}[t]
\centering\includegraphics[width=\textwidth]{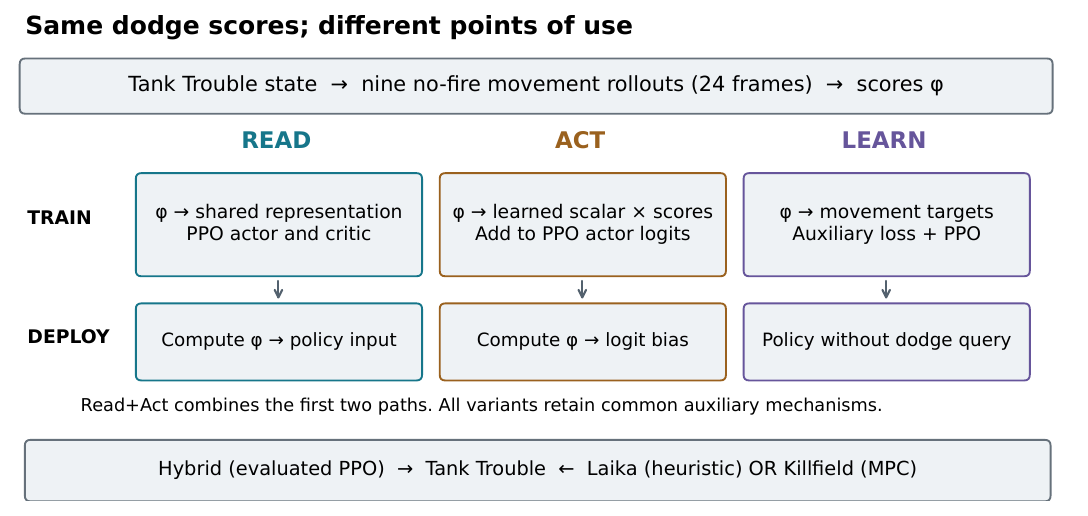}
\caption{One dodge signal, three insertion points. Read and Act participate in both
training and deployment; Learn uses the signal to construct training targets. Only
the additional dodge query is absent from Learn's policy requirements: common
auxiliary mechanisms remain. Read+Act combines the first two paths. Hybrid is the
policy varied across experiments; Laika and Killfield are fixed opponents.}
\label{fig:routes}
\end{figure*}

We ask: \emph{how does the use of the same action-conditioned lookahead signal affect
an existing learned game policy?} Our setting is deliberately specific: continuation
training of one policy trained with Proximal Policy Optimization (PPO)~\cite{ppo}
and augmented with auxiliary mechanisms in Tank Trouble. We compare shared input access,
a learned scalar logit bias, their combination, and an auxiliary-supervision
implementation. Figure~\ref{fig:routes} separates these insertion points and their
deployment requirements. The routes are composable design choices, not an exhaustive
or mutually exclusive taxonomy.

Our contributions are (1) a matched continuation study using the same dodge signal
and common ancestor; (2) a separation of endpoint performance from local sensitivity
to an added action path; and (3) explicit reporting of deployment dependencies and
the limits of available cost evidence. The strongest result concerns shared
input access. The auxiliary objective is an exploratory implementation, and we do
not infer a universally preferred route from this single environment.

\section{Design Space and Related Work}
Two axes organize this study: \emph{where} the signal enters computation and
\emph{whether} it remains required online. Imagination-Augmented Agents encode
imagined trajectories and combine them with model-free representations
\cite{i2a}, providing a precedent for learning to interpret predictions.
Soft Action Priors adapt the use of imperfect teacher information through reward
bonuses and auxiliary regularization \cite{sap}; they are related to our supervision
question, not an implementation of our scalar logit bias. Policy distillation
research likewise studies how learning objectives affect transfer \cite{distill}.
Residual Policy Learning combines a controller with a learned correction
\cite{rpl}; its continuous-action construction is a structural analogy rather than
the categorical-logit method evaluated here.

Mechanism comparisons also exist. \emph{The Mechanism Matters} varies knowledge-graph (KG) features, masking, shaping, and knowledge quality \cite{mechanism}. Our question
differs in its physical action-conditioned signal, pretrained policy, and potentially
redundant input and logit paths. In particular, our shuffled-path diagnostic retains
correctly ordered scores in the shared representation; it does not corrupt all
available knowledge.

Masking and shielding alter the permitted action set; bounded-prescience shielding
uses forward exploration to enforce a safety criterion \cite{bps}. Potential-based
reward shaping modifies learning rewards \cite{shaping}, while synthetic experience
and full planning change the data-generation or decision procedure. These are
relevant alternatives outside our implemented comparison. This is a scope decision,
not a claim that they cannot be compared fairly. We fix the environment reward,
action space, and interaction budget while intentionally varying parameterization
or auxiliary loss.

\section{Task, Agents, and Lookahead}
\subsection{Environment and agents}
\textbf{Tank Trouble} is the game environment, implemented by the project's duel
simulator: two tanks move through random mazes and fire ricocheting projectiles.
\textbf{Hybrid} is the evaluated learned policy, a PPO actor--critic with shared
map, bullet, and scalar encoders. Its existing aim/ballistic information,
ammo-related logit biases, and style losses are retained across variants. Base
therefore means \emph{no added dodge route}, not bare PPO without model assistance.

\textbf{Laika} is a fixed heuristic opponent. It selects prioritized goals such as
dodging, pursuing a firing opportunity, retreating when out of ammunition, and
escaping walls. \textbf{Killfield} is the author's fixed MPC/online planner. It
simulates candidate controls with a hand-tuned objective and short movement
commitments. Against Hybrid it holds the opponent's current controls in its
prediction model rather than simulating Laika's script. Neither opponent learns
during these runs. Appendix~\ref{app:agents} summarizes their decision procedures
and Hybrid's architecture. Using both tests whether route differences persist under two
different opponent behaviours; it does not compare PPO, rules, and MPC as competing
learning algorithms.

The simulation advances at 25 Hz. Hybrid chooses among nine movement controls
(three throttle settings by three turn settings), each with a fire/no-fire variant:
18 categorical actions. The maximum nominal round length is 750 frames, with a
post-kill settlement window for in-flight projectiles. Mutual destruction and
surviving victory are different outcomes. The fixed observation allocation has
1,028 components, including nine dodge slots; routes without input access zero
those slots. Reward and network details appear in Appendix~\ref{app:protocol}.

\subsection{The dodge signal}
Let $m\in\{0,\ldots,8\}$ index a movement and $f\in\{0,1\}$ the fire bit. For
each movement, a no-fire rollout holds that control for $H=24$ frames (0.96 s).
The resulting score $\phi_m(s)\in[-1,1]$ summarizes predicted short-term safety,
including death and clearance information. If a candidate survives the rollout,
its score is $\min(d_{\min}/8,1)$, where $d_{\min}$ is the minimum tank--bullet
centre distance in maze-cell units, initialized at eight. If it dies after $e$
completed rollout steps, its score is $-1+0.5e/H$. Rollouts can stop earlier when
the round freezes or ends. An early exit returns all ones if no active bullet is
present and the opponent's fire input is false. These scores are
not calibrated survival probabilities, long-term action values, or Killfield's
complete planning scores. Figure~\ref{fig:dodge_example} in Appendix~\ref{app:protocol}
shows this scoring on one recorded state.

Exact simulator physics does not make the future opponent behaviour exact: the
lookahead uses a control-holding approximation. The same movement score is copied
to both fire variants in Act, although the candidate simulation itself does not
fire. This is an inductive bias, not a guarantee that actual firing and non-firing
outcomes are equally safe.

\subsection{Observation timing}
Native evaluation publishes observations after applying controls and advancing the
game: $s_{t+1}=F(s_t,a_t,u_t)$, then $o_{t+1}=O(s_{t+1})$. Thus projectiles from
accepted shots are already part of the observed state before the next Hybrid
choice. Observing such an event is not observing a future action. The rollout also
copies simulator control fields; raw fire input is not identical to an accepted
shot, because ammunition and trigger state affect firing. We therefore describe
the signal as simulator-assisted and do not assert that every input it uses is
reconstructible from the public observation history. The web viewer's asynchronous
reaction-delay control is separate from this synchronous evaluation protocol.

\section{Compared Routes and Protocol}
\subsection{Runtime input and action paths}
Write $h_\theta(o,v\phi)$ for shared features, with $v\in\{0,1\}$ controlling
input access. Let $c_{m,f}$ contain the common existing biases. The logits are
\begin{equation}
 z_{m,f}=w_{m,f}^{\mathsf T}h_\theta(o,v\phi)+c_{m,f}
             +\alpha(P\phi)_m .
\label{eq:routes}
\end{equation}
Read enables $v$; Act enables the final term with a learned scalar $\alpha$ shared
across states. It is initialized to zero, but is not fixed throughout training.
Read+Act enables both. $P$ is the identity except for the shuffled-path control,
which uses a fixed permutation only in the last term. The shared input remains
ordered. The critic reads the same shared features, so Read does not isolate actor
learning from value learning.

\begin{table}[!ht]
\centering\small
\begin{tabular}{lcccc}
\toprule
Route & Input & Logits & Aux. loss & Online $\phi$\\
\midrule
Base & -- & -- & -- & no\\
Read & yes & -- & -- & yes\\
Act & -- & yes & -- & yes\\
Read+Act & yes & yes & -- & yes\\
Learn & -- & -- & yes & no\\
\bottomrule
\end{tabular}
\caption{Use of the additional dodge signal. Each listed implementation has three
completed continuation seeds and endpoint evaluations in the frozen snapshot.
Learn's objective was selected in a smoke test and is exploratory. The shuffled
runtime path is a diagnostic variant of Read+Act, and shuffled Learn a diagnostic
variant of Learn; neither is a separate integration route.}
\label{tab:routes}
\end{table}

\subsection{Training-only supervision}
Learn uses the Base deployment architecture and adds a movement-target loss during
training. For state $i$, define
\begin{align}
q_{im}&=\operatorname{softmax}(\phi_i/\tau)_m,\\
p_{im}&=\sum_{f=0}^{1}\pi(m,f\mid o_i),\\
w_i&=\operatorname{std}_m(\phi_{im}),\quad
\ell_i=-\sum_mq_{im}\log(p_{im}+10^{-8}).
\end{align}
The implementation normalizes the dispersion-weighted loss over the minibatch:
\begin{equation}
\mathcal L_\phi=\frac{\sum_i w_i\ell_i}{\max(\sum_iw_i,10^{-6})},\qquad
\mathcal L=\mathcal L_{\rm common}+\lambda\mathcal L_\phi.
\end{equation}
Equal-score states have zero weight. We use $\lambda=0.01$ and $\tau=0.3$, selected
from one smoke test, not a tuning sweep. The student is encouraged to match a
short-horizon safety heuristic, not an expert game-winning policy. Learn's network
architecture matches Base; its trained weights need not. It requires no deployment
dodge query, but this architectural property is distinct from whether the shared
environment implementation still computes unused scores.

\subsection{Matched continuation and evaluation}
All listed routes start from the same checkpoint with 404,881,408 cumulative
training steps and receive 4,194,304 further steps, ending at 409,075,712.
Continuation seeds are 21, 22, and 23; these are not three independent pretraining
histories. The runs use 256 environments and a fixed 50/50 Laika/Killfield mixture,
without frozen opponents during this continuation. The common backbone, reward,
style losses, and step budget are unchanged. New dodge input columns and the
scalar path start with zero influence in the original runtime comparison.
Learn additionally changes the loss, as specified above. Equal environment steps
do not imply equal optimization work or wall-clock cost.

Endpoint evaluation uses the final greedy policy: 2,000 episodes against Laika and
1,000 against Killfield per seed. We use saved reports with source hashes in the
accompanying aggregate data. Two frozen
policies excluded from this continuation mixture provide a limited held-out check.

We report per-seed results and descriptive intervals on paired seed differences
using Student's $t$ with two degrees of freedom. Pairing is by continuation seed,
not by individual evaluation episode. The underlying saved percentages are rounded
to one decimal place, so the intervals are approximate. Three seeds leave substantial
uncertainty; lack of a detectable difference does not establish equivalence.
These are screening and exploratory analyses, not a retrospectively confirmed
preregistration. Appendix~\ref{app:protocol} records additional scope limits.

\FloatBarrier
\section{Results}
\subsection{Shared input yields the largest observed gain}
\begin{figure*}[t]
\centering\includegraphics[width=\textwidth]{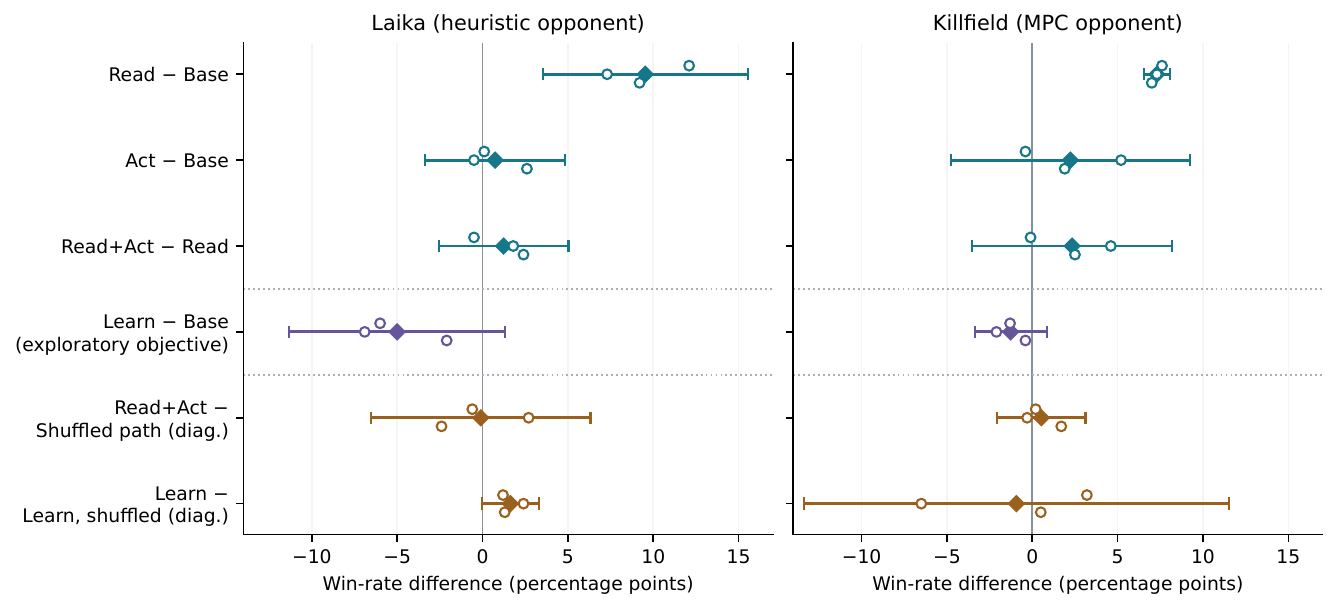}
\caption{Endpoint contrasts. Open circles show the three paired continuation-seed
differences; diamonds and bars show their mean and descriptive 95\% $t$ interval
($n=3$, df=2). Learn has three evaluated seeds, but its particular unswept auxiliary
objective remains exploratory. The bottom two rows (separated by a dotted line) are
the aligned/shuffled correspondence diagnostics for Act and Learn respectively, not
additional integration routes; both are compatible with a near-zero effect of
correspondence, with the Learn diagnostic's interval wide enough that it cannot rule
much out. The intervals describe variation across these continuations, not across
independently pretrained ancestors.}
\label{fig:contrasts}
\end{figure*}

Figure~\ref{fig:contrasts} compares incremental benefits rather than ranking five
mean bars. Read improves win rate over Base by 9.53 percentage points against Laika
and 7.30 against Killfield. Act alone gives smaller estimates, +0.73 and +2.23 points.
Adding Act to Read gives +1.23 and +2.33 points, with wide seed-level intervals.
Thus shared access gives the clearest positive result in this continuation setting;
we cannot exclude a useful additional action-path effect.

\paragraph{Training-only supervision does not recover the input gain.}
The three Learn runs yield 60.4/61.1/62.4\% against Laika and
51.7/46.8/51.5\% against Killfield: means of 61.3\% and 50.0\%.
Against the matching Base seeds, the mean differences are $-5.0$ and $-1.3$
percentage points. The Laika differences are negative in all three seeds;
Killfield differences are also negative, but smaller. This configuration does not establish a substitute
for online input access. It does not show that training-only supervision generally
fails: the teacher target, temperature, coefficient, and optimization interaction
remain specific to this implementation. Training entropy or switching changes alone
do not identify the reason for the endpoint result.

\paragraph{Training trajectories are a different measurement.}
The available training AUC integrates sampled-policy training win rates, not
periodic greedy evaluation. We therefore do not combine it with endpoint win rate
as evidence of a unified sample-efficiency ranking. The common step budget supports
a fixed-budget comparison; a learning-speed claim needs its own measurement.

\subsection{Online requirements differ; latency ranking is unresolved}
Read, Act, and their combination require computing the dodge scores online.
Learn and Base do not require this particular query, but retain common auxiliary
features. This difference concerns the policy's information requirements, not an
observed deployment speedup. The shared observation encoder used for evaluation
still computes the scores even for variants that ignore them. A deployment that
omits the unused query is therefore a separate implementation change.

We do not report a performance--latency frontier. A fair comparison would time the
same state workload, hardware, thread configuration, and inference batch size,
including observation construction and the verified omission of unused queries.
Equal environment-step budgets also do not imply equal training compute.

\subsection{Local sensitivity is small in an exploratory state bank}
A pathway can exist without substantially changing a decision. To measure local
influence, we evaluate identical states with identical checkpoint weights while
switching off only the actor's dodge term. Shared inputs and common biases remain
unchanged. The bank contains 9,216 states collected by Base, Read, and Read+Act
seed-21 policies (16 environments, 192 steps each, the same 50/50 opponent mixture).
It includes correlated frames and post-kill settlement, so it is an exploratory
sample rather than an independent set of decision trials.

\begin{figure*}[t]
\centering\includegraphics[width=\textwidth]{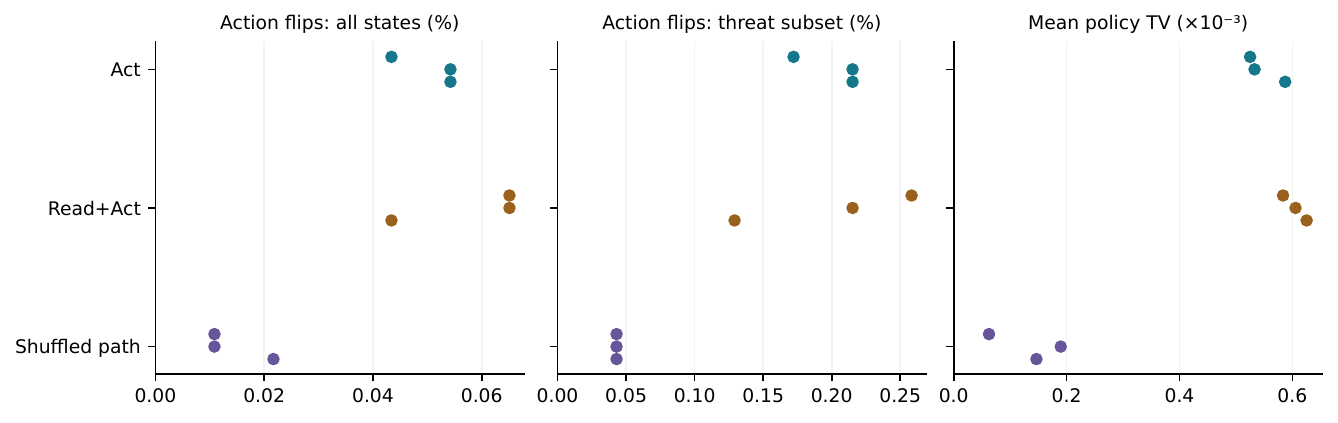}
\caption{Same-weight, same-state dodge removal. Each dot is one checkpoint, not an
independent state experiment. The first two panels measure changes in the greedy
18-action choice; the third measures mean total variation of the categorical
policy. Threat denotes $\min_m\phi_m<-0.3$ in this audit. These are local
interventions, not retraining or long-horizon performance evaluations.}
\label{fig:influence}
\end{figure*}

Across the Read+Act checkpoints, removal changes only 4--6 of the 9,216 greedy
actions, with mean policy TV between 0.000584 and 0.000625. Figure~\ref{fig:influence}
uses a single consistent intervention protocol for Act, Read+Act, and the shuffled
path. Nonconstant scores occur in 41.9\% of the bank; the threat subset is 25.2\%.
These state frequencies are specific to this audit, not a deployment-workload estimate.

Separately trained aligned and shuffled Read+Act variants differ in mean win rate
by $-0.10$ points against Laika and $+0.53$ against Killfield. Their descriptive
seed-level intervals are $[-6.53,6.33]$ and $[-2.05,3.12]$. This does not show
that correspondence is unnecessary: correctly ordered scores remain in the trunk,
and the shuffled scalar can approach zero. Small local sensitivity is compatible
with these results, but does not prove why independently trained models converge.

Applying the same aligned/shuffled diagnostic to Learn, the three-seed means differ
by $+1.63$ points against Laika and $-0.93$ against Killfield, with descriptive
seed-level intervals $[-0.02,3.29]$ and $[-13.37,11.51]$ (the wide latter interval
is driven by one seed's large swing). This does not identify an incorrectly ordered
teacher target as the source of Learn's endpoint gap; the coefficient and
temperature remain the more directly implicated, unswept design choices.

\subsection{The input advantage persists against two held-out policies}
Table~\ref{tab:heldout} evaluates two frozen policies from the project's later
production lineage, H1 and H2. They are absent from the continuation mixture, not
independent environments or agent families. Read retains a higher mean win rate
than Base against both. The precise Read versus Read+Act order is not stable, so
we describe persistence of the input advantage rather than a general route ranking.

\begin{table}[!ht]
\centering\small
\begin{tabular}{lrr}\toprule
Route & H1 & H2\\\midrule
Base & 13.6 & 11.6 \\
Read & 20.4 & 18.1 \\
Act & 14.2 & 13.1 \\
Read+Act & 20.8 & 17.5 \\
Learn & 12.5 & 9.2 \\
\bottomrule\end{tabular}
\caption{Win-rate means (\%) against two frozen policies excluded from continuation
training. Three seeds per row. Runtime-arm reports use approximately 1,000 episodes
for seed 21 and 600 for seeds 22/23; Learn uses 600 per seed. Unequal episode counts
and rounded reports make this a descriptive transfer check, not episode-paired
inference. Exact reported counts are supplied in the ancillary data.}
\label{tab:heldout}
\end{table}

\FloatBarrier
\section{Discussion and Limitations}
\paragraph{Practical interpretation.}
When this dodge computation is available, shared input access is the first route
supported by a clear positive result in our setting. A separate action path should
be assessed by what it changes, not just its existence or final coefficient. If
deployment cannot accommodate an additional dodge query, training-only supervision
is a relevant design option, but our evaluated objective does not yet recover the
input benefit. These are conditional lessons from this case, not a route-selection
rule for all game AI systems.

\paragraph{Why an additive path can have little immediate effect.}
For frozen logits $z$, let $k$ be the unique greedy action and $\delta$ the added
bias. The action remains uniquely maximal exactly when
$z_k-z_b>\delta_b-\delta_k$ for every $b\ne k$. Thus a minimum actor margin larger
than the bias span is sufficient to prevent a greedy flip. This statement alone
says nothing about distributional change, which is why we also measure TV.
Saturation, redundant access through the trunk, and disagreement between short-term
safety and game-winning actions are candidate explanations; the current experiments
do not identify their causal contributions. Hard masks, unlike finite biases, can
remove actions from the policy support.

\paragraph{Scope of inference.}
The three seeds share one mature ancestor, one environment, and exact simulator
physics with approximate opponent continuation. They do not establish effects of
policy maturity, independent initialization, imperfect dynamics models, or transfer
to games such as MOBAs. Read jointly affects actor and critic learning. Learn
changes the training objective as well as deployment requirements. The present data
do not quantify matched online latency or total training cost. Our observations
motivate separately measuring information access, local influence, and task benefit;
they do not establish a new general theory or formal safety guarantee.

\paragraph{Exploratory production changes are not route controls.}
Later production experiments introduced state-dependent gates and more self-play,
but lack matched continuation controls. An existing removal experiment also disabled
both dodge and ammo paths together. Neither identifies the isolated value of a
dodge gate or a training-scale effect, so neither enters our main route comparison.

\section{Conclusion}
In this Tank Trouble continuation study, exposing a short-horizon dodge signal to
Hybrid's shared actor--critic produces the largest observed gain. A learned scalar
action bias adds a smaller, uncertain increment and exhibits weak local influence
in an exploratory state bank. A three-seed auxiliary-supervision implementation
does not recover the input benefit, although it removes the policy's requirement
for a deployment dodge query. The findings support distinguishing a signal's
availability, decision-level influence, and task benefit while keeping the limits
of the deployment-cost evidence explicit.

\FloatBarrier
\appendix
\section{Protocol and Implementation Details}
\label{app:protocol}
All routes use the common ancestor and continuation budgets specified in
Section~4.3. Unused dodge input columns remain allocated but are zeroed before
encoding. The runtime shuffled-path control permutes the nine scores as
$(4,7,1,8,3,0,2,5,6)$, using zero-indexed movements; the shared input is unchanged.

\begin{center}
\includegraphics[width=0.8\linewidth]{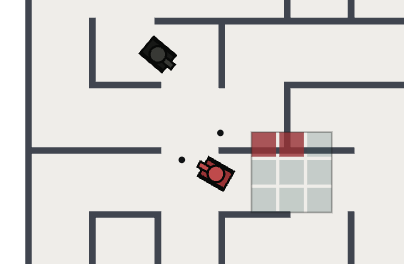}
\captionof{figure}{One recorded state, rendered with a separate visualization script
(not the interactive web viewer described below). The overlaid $3\times3$ grid
enumerates the nine no-fire throttle$\times$turn candidates in the same layout used
to compute $\phi$ (Section~3.2); rows top-to-bottom are forward/neutral/back and
columns left-to-right are left/none/right. The nine $\phi$ values for this state are
$\begin{smallmatrix}-0.958&-0.958&0.036\\0.055&0.055&0.055\\0.053&0.062&0.067\end{smallmatrix}$;
a cell is shaded red because $\phi<0$ there (a candidate that dies within the rollout,
both red cells equalling $-1+0.5\cdot2/24$, i.e.\ death after two rollout steps) and
grey because $\phi\geq0$ (the candidate survives, per $\min(d_{\min}/8,1)$). This
illustrates the scoring for one state; it is not a claim about which action any
route actually selects there.}
\label{fig:dodge_example}
\end{center}

\paragraph{Reward and common losses.}
A surviving win receives 1.0 within 10 seconds, logarithmically declining to 0.5 at
30 seconds. Mutual destruction gives $-0.1$; loss and timeout give $-1$.
A terminal movement-style bonus is at most 0.25: full credit for action-change rate
at or below 13\%, logarithmically declining to zero at a change every frame.
The PPO objective retains persistence and wall-contact losses with coefficients
0.02 and 0.05. All arms retain the common ammo-bias configuration.

\paragraph{Optimization.}
Runs use 256 environments, 128-step rollouts, four epochs and eight minibatches per
update, $\gamma=0.999$, GAE $\lambda=0.95$, PPO clip 0.2, value coefficient 0.5,
and gradient clipping at 0.5. The configured learning rate is $3\times10^{-4}$ with
the existing cumulative-step schedule (600M-step horizon), not a fresh constant
rate. There are 20 critic warmup updates; auxiliary supervision is inactive during
critic-only updates. These hyperparameters describe the continuation stage, not the ancestor's full training history.

\paragraph{Evaluation and provenance.}
The ancillary data include rounded per-seed endpoint results, held-out results
with their actual episode counts, and same-state diagnostic summaries. H1 and H2
are fixed checkpoints from later stages of the same production lineage, excluded
from the continuation training mixture. Their shared lineage limits the scope of
this transfer test. Each row averages the three continuation seeds equally.

Evaluation does not establish identical opponent sampling streams or episode-level
pairing. Screening-plan decisions and analyses were revised during development;
we report exploratory, descriptive results rather than retrospective confirmation.
The data accompanying this version support reconstruction of the aggregate plots;
simulator code, full training histories, and model checkpoints are not included in
this submission. This limits independent end-to-end replication.

\begin{table}[!ht]
\centering\footnotesize
\begin{tabular}{lrr}
\toprule
Route & Laika & Killfield\\
& Seeds 21 / 22 / 23 & Seeds 21 / 22 / 23\\\midrule
Base & 66.4 / 68.0 / 64.5 & 53.0 / 48.9 / 51.9 \\
Read & 78.5 / 75.3 / 73.7 & 60.6 / 56.2 / 58.9 \\
Act & 66.5 / 67.5 / 67.1 & 52.6 / 54.1 / 53.8 \\
Read+Act & 78.0 / 77.1 / 76.1 & 60.5 / 60.8 / 61.4 \\
Shuffled path & 78.6 / 74.4 / 78.5 & 60.3 / 61.1 / 59.7 \\
Learn & 60.4 / 61.1 / 62.4 & 51.7 / 46.8 / 51.5 \\
Learn (shuffled) & 59.2 / 58.7 / 61.1 & 48.5 / 53.3 / 51.0 \\
\bottomrule
\end{tabular}
\caption{Frozen endpoint reports (win \%). Three continuation seeds per arm;
2,000 Laika and 1,000 Killfield games per seed. Shuffled path permutes the
runtime bias in Read+Act; Learn (shuffled) permutes the auxiliary training target
in Learn. The two diagnostics permute different quantities and are not directly
comparable to each other.}
\label{tab:endpoints}
\end{table}

\section{Diagnostic Protocol}
The same-state audit uses collection seeds 707000--707002, with Base, Read, and
Read+Act seed-21 collectors. The ancillary diagnostic summary gives each evaluated
checkpoint's scalar coefficient, action-flip fraction, threat-conditioned flip
fraction, and mean policy TV. Greedy flips refer to the full 18-action choice.
Scores in the shared representation and all common biases are held fixed during
removal. The states include settlement frames and are correlated within trajectories;
we therefore do not attach independent-trial confidence intervals to the flip rates.
No probe establishes that a path was unused throughout training or has no long-term
return effect.

\section{Agent Decision Procedures}
\label{app:agents}
Tank Trouble is the environment. Hybrid is the learned policy whose lookahead
routes are varied; Laika and Killfield are fixed opponents. This appendix explains
how each agent chooses actions, rather than treating them as competing ablation arms.

\subsection{Hybrid: learned actor--critic with auxiliary mechanisms}
Hybrid combines three observation encoders: a CNN for the maze grid, a shared MLP
for bullet features with masked mean/max pooling, and a scalar encoder for tank
state and auxiliary information. Their outputs feed a 256-unit shared trunk.
An actor head produces 18 action logits (nine throttle/turn combinations, each
with or without firing), while a critic head estimates state value for PPO training.
The allocated network has approximately 229k parameters. Training samples actions;
reported endpoint evaluation takes the largest logit.

Common aim/ballistic features, ammo-related action biases, and persistence and
wall-contact losses remain enabled. The controlled variation is specifically how
the additional nine dodge scores enter this system: shared features, an additive
actor bias, or a training target. The critic shares the feature representation;
it does not directly choose actions. Hybrid is therefore a neural decision-maker
with auxiliary mechanisms, not a wrapper that simply executes the MPC opponent's
chosen action.

\subsection{Laika: prioritized heuristic goals}
Laika maintains a current goal and repeatedly evaluates hand-coded alternatives.
Incoming projectile threats favour dodging; reachable opponents favour pursuit and
shooting; depleted ammunition favours retreat; wall contact can trigger an escape
manoeuvre. Goal priorities arbitrate between alternatives, and the current goal's
priority decays so that another goal can take over. Pathfinding and geometric
projectile prediction help execute the selected goal.

Laika has no learned policy parameters and is not updated during training. Its
use of geometry and trajectory checks should not be confused with an absence of
prediction: what makes it heuristic is the manually specified goal-selection and
control logic, rather than a learned objective or optimization over full candidate
plans.

\subsection{Killfield: receding-horizon online planning}
Killfield evaluates candidate controls by simulating them forward in a physics
sandbox over a 36-frame horizon. Its hand-tuned scoring combines tactical progress,
firing opportunities, predicted outcomes, displacement, and wall-contact costs.
It selects a high-scoring plan and uses short control commitments (four frames for
movement and two for turning) to reduce rapid alternation between similar choices;
urgent events can trigger replanning.

Against Hybrid, its L1 opponent model holds the opponent's current controls during
prediction rather than invoking Laika's actual decision script. Planning constants
and scoring weights are fixed across the experiments. Although the planner and the
dodge signal share simulation infrastructure, Killfield's multi-term action search
is distinct from the nine no-fire, 24-frame safety summaries given to Hybrid.

The two opponents provide different behavioural tests for the same learned-policy
variants. Neither is trained alongside Hybrid in this comparison, and neither's
standalone strength is the paper's main research question.

\end{document}